# Surgical Phase Recognition of Short Video Shots Based on Temporal Modeling of Deep Features

Constantinos Loukas
*Laboratory of Medical Physics, Medical School*
*National and Kapodistrian University of Athens*
Athens, Greece
cloukas@med.uoa.gr

*Abstract*—Recognizing the phases of a laparoscopic surgery (LS) operation form its video constitutes a fundamental step for efficient content representation, indexing and retrieval in surgical video databases. In the literature, most techniques focus on phase segmentation of the entire LS video using hand-crafted visual features, instrument usage signals, and recently convolutional neural networks (CNNs). In this paper we address the problem of phase recognition of short video shots (10s) of the operation, without utilizing information about the preceding/forthcoming video frames, their phase labels or the instruments used. We investigate four state-of-the-art CNN architectures (Alexnet, VGG19, GoogleNet, and ResNet101), for feature extraction via transfer learning. Visual saliency was employed for selecting the most informative region of the image as input to the CNN. Video shot representation was based on two temporal pooling mechanisms. Most importantly, we investigate the role of 'elapsed time' (from the beginning of the operation), and we show that inclusion of this feature can increase performance dramatically (69% *vs.* 75% mean accuracy). Finally, a long short-term memory (LSTM) network was trained for video shot classification based on the fusion of CNN features with 'elapsed time', increasing the accuracy to 86%. Our results highlight the prominent role of visual saliency, long-range temporal recursion and 'elapsed time' (a feature so far ignored), for surgical phase recognition.

## I. INTRODUCTION

Laparoscopic surgery (LS), a common type of minimally invasive surgery (MIS), provides not only substantial therapeutic benefits for the patient, but also the opportunity to record the video of the operation for reasons such as documentation, technique evaluation, skills assessment, and cognitive training of junior surgeons [1],[2]. However, a major technological challenge is the effective content management of the recorded videos, given that an operation may last for more than an hour, whereas the duration of yearly operations per surgeon may exceed 1,000 hours [3].

The traditional way of classifying/retrieving videos relevant to a particular feature of the operation is via text mining from manual annotations. Apart from the operation type, the annotation may include keywords such as a special technique performed, anatomic characteristics, or instruments utilized. However, this type of labelling has some limitations, preventing the effective management, representation and indexing of the recorded videos. First, manual annotation is tedious and time-consuming. Second, semantic characteristics that are discovered to be of the surgeon's interest at a later stage, are excluded from future searches. Third, global annotation of terms provides limited information about their time-stamp, unless this is manually inserted. In most cases, the surgeon performs manual skimming of the video to locate the object/event of interest, which is inefficient. In order to provide surgeons with additional tools for video content management, an effective way for video content representation is essential.

Automated surgical phase recognition from the LS video is an important topic of research, usually defined as 'surgical workflow analysis' (SWA). The phases of a surgical operation constitute fundamental temporal units, where the surgeon attempts to complete an overall task before advancing to the next phase. During the phase, the surgeon manipulates certain anatomic tissues with the surgical instruments, some of which may be specific to its label. Surgical phases are of crucial importance for the structure of the operation as they correspond to the top hierarchy level according to the 'operation decomposition' model: phases, steps, tasks/events, and gestures, as described in a recent review [4]. Hence, a method able to recognize the phase of a surgical operation would offer solutions to various challenges encountered in surgical video content management systems. Other applications include phase-based skills assessment, automatic selection of didactic content, and improved OR scheduling (if phase recognition is performed online).

Initial SWA works employed tool usage signals from RFID and electromagnetic (EM) sensors as well as manual annotations, based on the hypothesis that a surgical phase is characterized by a certain hand gesture or/and tool usage pattern [5],[6],[7]. However, employment of additional sensors may interfere with the operational workflow and there are concerns whether these data can be automatically acquired in the operating room.

Visual features extracted from the recorded video of the operation seem a more straightforward option due to the endoscopic camera employed. Compared to endoscopic examinations, surgical videos present significant challenges such as presence of smoke (coagulation), heavy interaction with the operated organs (dissection/clipping/cutting), and frequent camera motion as well as tool insertion/removal. Prior works on vision-based phase recognition included hand-engineered features based on color, texture, intensity gradients, or combinations of them [8]. In [9], gradient magnitudes, histograms and color values were employed. After dimensionality reduction based on tool usage signal data, laparoscopic cholecystectomy (LC) operations were segmented into 14 phases with accuracy close to 77%. The combination of visual features with tool usage signals was also employed in [10]. In another study, phase border detection of LC videos was performed via image-based instrument recognition [11].

The aforementioned works employ handcrafted features, which are specifically designed to capture certain type of information ignoring other image characteristics. Recently, deep learning approaches (e.g. Convolutional Neural networks-CNNs) have shown promising results for phase recognition. For example, Twinanda *et al.* proposed the EndoNet architecture, a CNN based on the AlexNet architecture, which was fine-tuned on a dataset of 40 LC operations for the task of tool and phase recognition [12]. The precision of offline and online phase recognition was close to 85% and 74% respectively. In the recent M2CAI 2016 challenge for online phase recognition of LC operations, Jin *et al.* achieved a mean jaccard score of 78.2% (the challenge allowed a 10s margin in the predictions). Their method combined feature vectors extracted from a fine-tuned ResNet50 CNN architecture combined with a long-short term memory (LSTM) network to encode temporal information [13]. These works focus on the online segmentation of the entire operation based on fully supervised training and using visual information from the preceding video frames as well as their inferred phase labels. Recently, a CNN-based method for video shot classification of laparoscopic gynecologic actions was proposed in [14]. Video shots of the entire course of each surgical action were employed for training and testing.

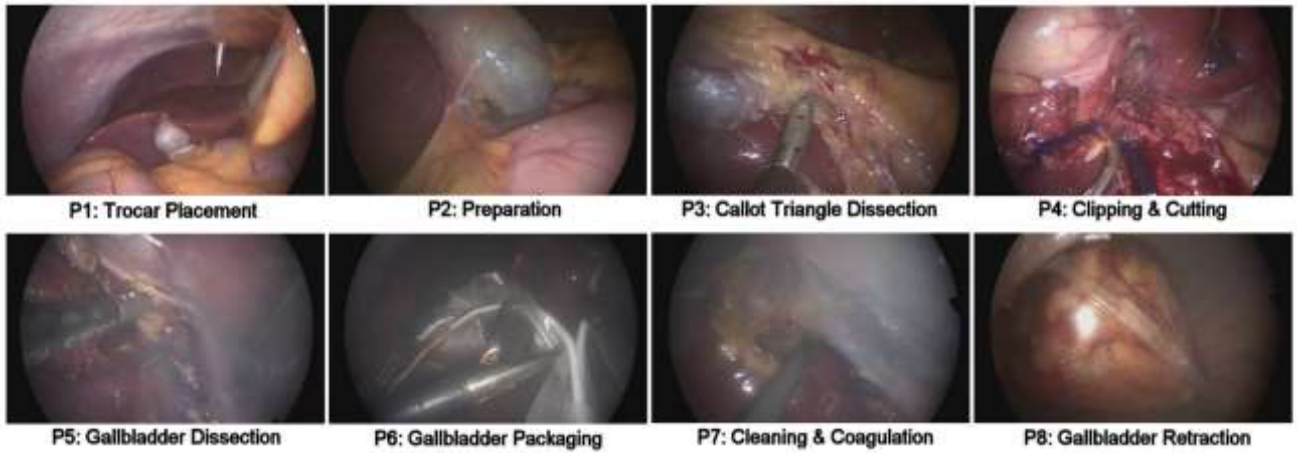

Fig. 1. An overview of the 8 surgical phases (P1-P8). The frames were extracted from the annotations of a single LC operation.

In this paper we present a method for phase recognition from short video shots (10s) of surgical operations, without any prior knowledge about the preceding/forthcoming video frames, phase labels or instruments used. We concentrated on LC which is a fundamental operation for junior surgeons. Each video shot essentially represented only a small fraction of the entire phase. Prompted by the advances in image classification based on deep learning approaches, we investigated four state-of-the-art CNN architectures (Alexnet, VGG19, GoogleNet, and ResNet101), for feature extraction via transfer learning. Features were extracted from two different types of receptive fields: one based on traditional frame resize to match the CNN's input size and another one based on the most salient region of the input image. Video shot representation was performed via two temporal pooling mechanisms. Initially, video shot classification was based on the 1$^{st}$ nearest-neighbor (NN) using two different distance metrics. Most importantly, we investigated the role of absolute 'elapsed time' (from the beginning of the operation), in video shot classification and we present results that show that the inclusion of this feature can increase performance dramatically. Finally, we fused the CNN features with 'elapsed time' and applied long-range temporal recursion to estimate the probability of each surgical phase for a video shot, which improved even further the classification performance.

## II. METHODOLOGY

### A. Video Shot Dataset

In this work we employed surgical videos from the M2CAI 2016 Challenge Dataset, which includes video recordings of complete LC operations [12],[15]. In particular, we analyzed 27 video recordings (about 19 hours total duration), from the 'workflow-train' sub-dataset. The videos are recorded at 25 frames per second (25 fps), with full HD resolution 1920×1080. Each video includes frame-by-frame annotations of the 8 surgical phases of LC: P1) Trocar Placement, P2) Preparation, P3) Calot Triangle Dissection, P4) Clipping & Cutting, P5) Gallbladder Dissection, P6) Gallbladder Packaging, P7) Cleaning & Coagulation, and P8) Gallbladder Retraction. Sample frames of the surgical phases are presented in Fig. 1. It may be seen that some phases may present distinct image characteristics (e.g. P1, P8), whereas others (e.g. P5-P7) are quite challenging for visual recognition due to their similarity or/and presence of smoke (e.g. see P5, P6, P7).

It should be emphasized that not all operational phases are sequential and they are governed by some temporal constraints: P6, P7, P8 are not always sequential (P6 may occur after P7, or/and P7 after P8), P7 may occur 2 times, and P7 may not be present at all. However, phases P1-P5 are always present in an operation and occur sequentially. A statistical overview of the phases is presented in Table 1. It may be seen that P2 and P6 have the shortest duration whereas P3 and P5 have the longest one.

TABLE I. STATISTICAL OVERVIEW OF THE PHASES

| Phase ID | mean ± std (minutes) | min (minutes) | max (minutes) |
|---|---|---|---|
| **P1** | 3.04 ± 1.70 | 1.42 | 7.08 |
| **P2** | 1.71 ± 2.06 | 0.35 | 11.03 |
| **P3** | 10.53 ± 7.97 | 1.95 | 26.82 |
| **P4** | 4.70 ± 2.87 | 0.94 | 12.39 |
| **P5** | 10.40 ± 6.30 | 1.68 | 24.11 |
| **P6** | 1.14 ± 0.59 | 0.28 | 3.03 |
| **P7** | 5.68 ± 2.80 | 1.01 | 15.97 |
| **P8** | 4.93 ± 5.61 | 0.66 | 22.26 |

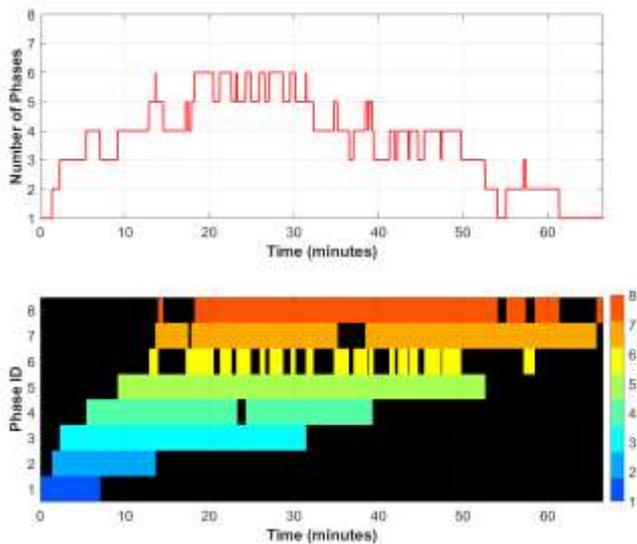

Fig. 2. Graphical overview of the number of simultaneously occurring phases (top) and the temporal span of each phase (bottom), across the dataset of 27 surgical operations.

Fig. 2 shows the number of simultaneously occurring phases as well as the timings of each phase, across all 27 video recordings, in the same diagram. From the top diagram it may be seen that up to the first 10 minutes, about 1-3 phases may occur (from P1-P4), and the same is valid after the 50th min (from P5-P8). From the bottom diagram it may be seen that the order of the phases is relatively fixed and their duration is limited (e.g. P1-P5). Furthermore, some phases present almost no temporal overlap (e.g. P1 *vs.* P4-P8 and P2 *vs.* P6-P8), whereas some others present moderate overlap (e.g. P1 *vs.* P3 and P2 *vs.* P5). From this analysis it is evident that the absolute temporal position of a video frame (with regard to the beginning of the operation), is a crucial factor that should be considered in phase recognition.

Based on the aforementioned dataset, we extracted the video shot dataset which was employed for classification. In particular, for each video and for each of the 8 phases, we extracted 2 non-overlapping shots of 10s duration (i.e. 250 frames). The video shots were extracted from random temporal positions of each phase, ensuring that the first/last frame of each shot was within the temporal limits of each phase. In two videos, P7 was absent so in order to have an equal number of shots per phase, we randomly selected 50 (out of 54) shots for each of the P1-P6 and P8 phases, leading to a total number of 400 video shots, equally distributed across the 8 phases.

*B. CNN Feature Extraction*

Feature extraction was based on transfer learning using 'off-the-shelf' features extracted from four state-of-the-art CNNs: Alexnet, VGG19, GoogleNet, and Resnet101. These network architectures were chosen as they are known to perform well on surgical endoscopy images [3]. Transfer learning implies that the CNNs were pretrained, in this case on the Imagenet database which contains millions of natural images distributed in 1000 classes. Although surgical images are substantially different, given the powerful architecture of the CNNs and the huge volume of Imagenet, transfer learning has been proved a simple, yet good-working approach for content-based description of surgical images [14]. Moreover, our dataset is considerably small to train these CNNs from scratch. However, as will be discussed later, we perform training to model the temporal variation of the extracted CNN features.

Alexnet consists of eight layers: 5 convolutional layers followed by 3 fully-connected (FC) layers. For each frame in a video shot, we extracted features from layer fc7, which is the before-final-fc (BFFC) layer with length $n_1$=4096. VGG19 is much deeper, consisting of 16 convolutional layers followed by 3 fully connected layers. We again extracted features from the BFFC layer (fc7, $n_2$=4096). GoogleNet is different to Alexnet and VGG19, including various Inception modules with dimensionality reduction and only one fully connected layer combined with a softmax layer (22 layers in total). For each frame we used the features extracted from the BFFC layer: pool5-7x7_s1 ($n_3$=1024). Finally, the Resnet101 model is the deepest of the four (101 layers); it stacks several residual blocks in-between the convolutional blocks aiming to alleviate the vanishing gradient problem, usually encountered when stacking several convolutional layers together. For the ResNet101 model we used the bottleneck features extracted from the BFFC layer: pool5 ($n_4$=2048).

Based on the aforementioned approach we extracted feature descriptors from each video frame. In order to achieve a compact feature representation of the video shot, we concatenate the descriptors along the temporal dimension and apply two temporal pooling mechanisms: max-pooling and average-pooling. The former extracts the maximum value from each dimension of the BFFC layer, whereas the second one outputs the average from each dimension of the BFFC layer. For each CNN architecture employed, both approaches result in a single feature descriptor for each shot, equal to the size of the corresponding BFFC layer.

*C. CNN Input and Saliency Maps*

The size of the input layer of the aforementioned CNNs is: 227×227 (Alexnet) and 224×224 (VGG19, GoogleNet, and ResNet101). In previous works, the original image is resized either to match the CNN's input, or so that the smaller side matches one side of the CNN input layer and then the center crop is used as input to the CNN [14],[16]. However, both approaches have some limitations. Considering that the original video resolution is 16/9, the former case leads to a spatial degradation of the original image, as the aspect ratio is forced to be 1 (see Fig. 3). In the latter case, image resizing does not affect the aspect ratio, but extracting features from the center crop may lead to an efficient representation of the original frame since the structures of interest are not always in the center (w.r.t. Fig. 1, in P1 the trocar is located towards the upper-right corner whereas in P4 the clips/tool-tip are in the bottom).

In this work, we propose an alternative mechanism for region selection based on visual saliency. Integration of visual saliency in CNN-based content-based image representation is a major trend nowadays. The main idea is to generate a saliency map that represents the most salient regions of the input image, without any prior assumption, based on various criteria such as color- and texture-contrast [17],[18]. Recent works have shown that using as input to the CNN a salient, instead of a center/resized crop, image provides better classification results [19]. In this work we have employed the static version of the adaptive whitening saliency (AWS) methodology, which has shown superior performance in predicting human attention [17]. Recently, this method was applied for keyframe extraction from video shots of LC operations [20]. In brief, the model employs a bank of 2D LogGabor filters, generating a series of filter response maps, which are then accumulated to generate the final saliency map of the image.

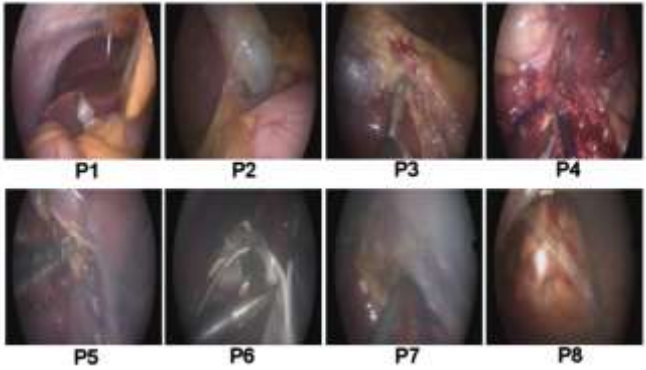

Fig. 3. The images shown in Fig. 1, resized to 224×224.

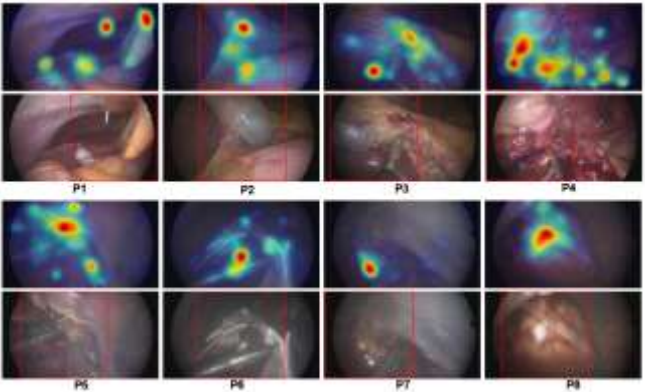

Fig. 4. The saliency maps overlaid on the images shown in Fig. 1 (top) and the rectangular patches which were used as input to the CNN (bottom).

For the purpose of this work, first we resized the original full HD frame of the video shot so that the smaller side (height) matched one side of the CNN input layer (i.e. 224 or 227). Second, we computed the saliency map of the resized frame based on the AWS model. Third, we computed the 5 strongest local maxima of the saliency map using a neighborhood size 9×9. Fourth, the spatial coordinates of these maxima were averaged providing the center location of an image patch that was used as input to the CNN. In case the center location was so that the patch lied outside the resized frame, the patch was shifted to lie within the frame window. Finally, a feature descriptor was extracted from the BFFC layer of the CNN, as described previously.

Fig. 4 provides the saliency maps overlaid on the corresponding resized frames shown in Fig. 1. Below are the rectangular patches (in this case 224×224), which were used as input to the CNN. It may be seen that most of the structures of interest lie within the image patch, such as the trocar in P1, the gallbladder in P2, the tools in P3-P7, and the retrieval bag in P8.

### D. Video Shot Classification Based on Temporal Pooling

Video shot classification was based on the $1^{st}$ NN using two different distance metrics: Euclidean and cosine. For both metrics we used the compact feature representation of each shot based on the max- and average-pooling mechanisms, described before. The Euclidean metric simply takes the L2 norm of the compact feature descriptors whereas the cosine distance is defined as one minus the cosine of the included angle between the descriptors. Each of the 400 compact feature descriptors of the corresponding video shots was treated as a candidate descriptor, for which the class is predicted based on its nearest distance among all other descriptors (treated as training descriptors). The results were obtained separately for each combination of pooling mechanism and distance metric.

As described in Section II.A, the time-stamp of the video frames seem to play an important role in the prediction of the surgical phase. Hence, we also investigated the effect of adding the time-stamp of the video shot as an additional dimension to the CNN features (n+1). Consequently, each video shot was represented by the pooled CNN descriptor concatenated by the time stamp of the shot (taken as the elapsed time, $T_{elapsed}$, of the 1st frame of the shot from the beginning of the operation).

### E. Video Shot Classification Based on Temporal Modeling

A significant limitation of the aforementioned pooling mechanisms for the video shot representation, is that the resulting descriptor does not take into account the temporal order of the individual frames. In other words, if one shuffles the frames of a video shot, then each of these mechanisms will lead to the same feature descriptor. However, the order of the frames (and so the extracted CNN descriptors) is an important piece of information that should be utilized in the classification task. A recurrent neural network (RNN) exhibits dynamic temporal behavior for a time sequence, as the connections between nodes form a directed graph along the sequence. Moreover, RNNs use their internal state to process input sequences and they are able to connect previous information to the present task, such as using previous video frames for the understanding of the present frame and eventually the class of the entire video shot.

In this work, the CNN feature extraction process was applied at each time point of the shot, leading to a temporal order of feature descriptors. Then, we employed the LSTM network, a special kind of RNNs, composed of a series LSTM units (as many as the number of shot's descriptors), and a FC layer with softmax at the end to perform classification into the eight phases. To training parameters of the LSTM network were similar to those proposed recently in [21]: batch size=16, training epochs=80, hidden units=200, learning rate=0.001, the Adam method for optimization and cross-entropy as loss function.

Similarly to the aforementioned idea of concatenating the compact CNN shot descriptor with the time-stamp of the video shot, here we employed as input to the LSTM network the CNN features (extracted from each frame of the shot), concatenated with the time-stamp of the corresponding video frame (dimensionality: n+1). The model was run on every 25th frame (i.e. 1 fps) to reduce the computational cost. During training/testing the temporal order of the feature descriptors was preserved. The LSTM network was trained on 50% of the video shot dataset (i.e. 25 video shots per class), and the remaining 50% served as the test set. We performed 5 random cycles of training, making sure that a video shot was included at least once in a training cycle, and then we averaged the evaluation metrics (see next section).

TABLE II. CLASSIFICATION RESULTS BASED ON RESIZED RAW IMAGES (FEATURES: CNN)

| CNN type | Pooling: Max Distance: Cosine | | | | Pooling: Average Distance: Cosine | | | | Pooling: Max Distance: Euclidean | | | | Pooling: Average Distance: Euclidean | | | |
|---|---|---|---|---|---|---|---|---|---|---|---|---|---|---|---|---|
| | *Acc* | *Pre* | *Rec* | *F1* | *Acc* | *Pre* | *Rec* | *F1* | *Acc* | *Pre* | *Rec* | *F1* | *Acc* | *Pre* | *Rec* | *F1* |
| alexnet | 0.60 | 0.60 | 0.61 | 0.61 | 0.59 | 0.60 | 0.59 | 0.60 | 0.59 | 0.60 | 0.60 | 0.60 | 0.56 | 0.58 | 0.57 | 0.57 |
| googlenet | 0.60 | 0.61 | 0.62 | 0.61 | 0.55 | 0.55 | 0.56 | 0.55 | 0.60 | 0.60 | 0.61 | 0.61 | 0.53 | 0.53 | 0.54 | 0.53 |
| vgg19 | 0.60 | 0.61 | 0.61 | 0.61 | 0.57 | 0.56 | 0.57 | 0.57 | 0.60 | 0.61 | 0.60 | 0.60 | 0.56 | 0.57 | 0.57 | 0.57 |
| resnet101 | **0.64** | **0.65** | **0.64** | **0.64** | 0.61 | 0.64 | 0.61 | 0.62 | **0.65** | **0.66** | **0.65** | **0.65** | 0.57 | 0.59 | 0.58 | 0.58 |

## III. EXPERIMENTAL RESULTS

The performance of the aforementioned approaches was evaluated in terms of the following metrics:

$$Acc = (TP+TN)/(P+N) \quad (1)$$

$$Pre = TP/(TP+FP) \quad (2)$$

$$Rec = TP/(TP+FN) \quad (3)$$

$$F1 = 2 \times Pre \times Rec/(Pre+Rec) \quad (4)$$

where: *Acc*, *Pre*, *Rec*, *F1* denote Accuracy, Precision, Recall, and F1-score, respectively; *TP*, *TN*, *FP*, *FN*, *P*, *N* denote: true positives, true negatives, false positives, false negatives, positives and negatives, respectively.

Table 2 shows average classification results using CNN features extracted from the resized raw images, for the two temporal pooling mechanisms and the two distance metrics. It is worth noting that for a particular CNN and pooling mechanism, the distance metrics are similar, except for ResNet101 where Euclidean leads to worse results for average pooling. With regard to the two pooling mechanisms, max pooling seems to yield better performance, especially when used with cosine, for all CNNs. The best performance was achieved by ResNet101 (~65%) using max-pooling with either distance metric.

Table 3 summarizes the classification results using features extracted from the most salient region of the image, as described in Section II.C. Average-pooling is omitted as it was proved to yield worse results. Compared to Table 2 it is clear that extracting features from the most salient image patch leads to 2-5% improvement for both distance metrics and for all CNNs. The cosine distance produced better results (by 3-5%), for all CNNs. The best performance was achieved again by ResNet101 (~70%), whereas the other three CNNs had similar performance (61-65%) although higher than that shown in Table 2.

Table 4 presents the results using the CNN features extracted from the salient patch, concatenated with the 'elapsed time' feature. Compared to Table 3, there is a notable improvement by ~5-10% for all metrics, CNN architectures, and distance metrics (except for GoogleNet/ResNet101 with Euclidean, which is the same). Note that the difference in this experiment was that the feature vector was increased by 1, the elapsed time of the 1st frame of the shot from the operation start. Again, the cosine distance produces better results than Euclidean, for all CNNs (~3-9% improvement). The best performance was achieved again by ResNet101, about 5% higher than that using only the CNN features (75% *vs*. 70%).

Table 5 provides the results based on LSTM model training. Similarly to Table 4, the input to the network was the CNN feature vector extracted from the most salient region of the video frame, concatenated with its time-stamp (i.e. elapsed time from operation start). Clearly the LSTM model yields superior performance across all metrics, compared to the naive NN cosine distance with max-pooling (compare to Table 4). Specifically, the improvement was about: 2%, 10%, 7% and 11% for Alexnet, GoogleNet, VGG19, and ResNet101, respectively. The best performance was achieved again by ResNet101 (86-88%), whereas the second best model was GoogleNet (~81%).

Table 6 illustrates the performance of the LSTM model for the individual classes, using a confusion matrix. Columns denote the predicted class while rows indicate the true class. The numbers denote the prediction percentage with respect to the samples from a particular class (positives). For phases P1-P4, the model yields almost perfect predictions, higher than 94%, and with very low or no confusion among the other classes. For P7 and P8 the results are also remarkable: 79% and 87% respectively. The model seems to slightly confuse P7 with P5 (9%), and much less with P4 and P8 (~5%). Phase P8 is slightly confused with P5 and P7 (5% and 8% respectively). For P5 and P6 the performance is similar (~72%). P5 is mostly confused with P7 (26%), whereas P6 with P7 (19%) and P4 (10%). The lower results for P5, P6 may be due to the visual similarity with P7 as a result of smoke, as depicted in Fig. 1.

TABLE III. CLASSIFICATION RESULTS BASED ON SALIENT MAPS (FEATURES: CNN)

| CNN type | Pooling: Max Distance: Cosine | | | | Pooling: Max Distance: Euclidean | | | |
|---|---|---|---|---|---|---|---|---|
| | *Acc* | *Pre* | *Rec* | *F1* | *Acc* | *Pre* | *Rec* | *F1* |
| alexnet | 0.62 | 0.61 | 0.62 | 0.61 | 0.57 | 0.57 | 0.57 | 0.57 |
| googlenet | 0.63 | 0.65 | 0.64 | 0.64 | 0.60 | 0.61 | 0.61 | 0.61 |
| vgg19 | 0.61 | 0.61 | 0.62 | 0.62 | 0.60 | 0.60 | 0.61 | 0.60 |
| resnet101 | **0.69** | **0.70** | **0.70** | **0.70** | 0.68 | 0.68 | 0.68 | 0.68 |

TABLE IV. CLASSIFICATION RESULTS BASED ON SALIENT MAPS (FEATURES: CNN AND $T_{ELAPSED}$)

| CNN type | Pooling: Max Distance: Cosine | | | | Pooling: Max Distance: Euclidean | | | |
|---|---|---|---|---|---|---|---|---|
| | *Acc* | *Pre* | *Rec* | *F1* | *Acc* | *Pre* | *Rec* | *F1* |
| alexnet | 0.71 | 0.73 | 0.71 | 0.71 | 0.68 | 0.68 | 0.68 | 0.68 |
| googlenet | 0.70 | 0.72 | 0.70 | 0.70 | 0.61 | 0.61 | 0.61 | 0.61 |
| vgg19 | 0.71 | 0.72 | 0.71 | 0.71 | 0.65 | 0.65 | 0.66 | 0.65 |
| resnet101 | **0.75** | **0.76** | **0.75** | **0.75** | 0.68 | 0.68 | 0.68 | 0.68 |

TABLE V. CLASSIFICATION RESULTS BASED ON SALIENT MAPS AND LSTM (FEATURES: CNN AND $T_{ELAPSED}$)

| CNN type | Acc | Pre | Rec | F1 |
|---|---|---|---|---|
| alexnet | 0.73 | 0.74 | 0.73 | 0.72 |
| googlenet | 0.81 | 0.82 | 0.81 | 0.81 |
| vgg19 | 0.78 | 0.77 | 0.78 | 0.77 |
| resnet101 | **0.86** | **0.88** | **0.86** | **0.86** |

TABLE VI.  CONFUSION MATRIX BASED ON SALIENT MAPS AND LSTM (FEATURES: CNN AND T$_{ELAPSED}$)

| True/Pred. | P1 | P2 | P3 | P4 | P5 | P6 | P7 | P8 |
|---|---|---|---|---|---|---|---|---|
| P1 | **1.00** | | | | | | | |
| P2 | | **0.94** | 0.06 | | | | | |
| P3 | | 0.05 | **0.94** | 0.01 | | | | |
| P4 | | | | **0.93** | 0.01 | 0.01 | 0.01 | 0.04 |
| P5 | | | | | **0.72** | | 0.26 | 0.02 |
| P6 | | | | 0.10 | | **0.71** | 0.19 | |
| P7 | | | | 0.04 | 0.09 | 0.02 | **0.79** | 0.06 |
| P8 | | | | | | 0.05 | 0.08 | **0.87** |

## IV. CONCLUSIONS

In this paper we propose a method for video shot classification into surgical phases based on deep features and temporal information modeling. Our results lead to the following conclusions. First, extracting CNN features from the most salient regions of the image allows to achieve better results (up to 5%). Second, when using a NN approach for classification, the cosine distance provides better results (up to 5%). Third, video shot representation based on max-pooling of CNN image features is better than average pooling (up to 6%). Fourth, deeper CNNs provide more robust features for classification (up to 10% improvement). Fifth, 'elapsed time' (a feature missing from the related literature), can increase performance dramatically (up to 10% and 6% for shallower and deeper architectures, respectively). Finally, employing an LSTM model for temporal modeling of the CNN features fused with 'elapsed time' provides significant performance improvement: 86% accuracy and 88% precision (compared to 75% and 76% when max-pooling is employed, respectively). The investigation of a visual saliency model specialized to surgical videos, fine tuning of a ResNet model in which 'elapsed time' is embedded, and other temporal information modeling architectures, are major topics of interest for future research work.